\pdfoutput=1

\documentclass[11pt]{article}

\usepackage[final]{coling}
\usepackage{amsmath}
\usepackage{times}
\usepackage{latexsym}

\usepackage[T1]{fontenc}

\usepackage[utf8]{inputenc}

\usepackage{microtype}

\usepackage{inconsolata}

\usepackage{graphicx}

\usepackage{framed}
\usepackage{enumitem}
\usepackage{tabularx}
\usepackage{tabulary} 
\usepackage{subcaption}
\usepackage{tikz}
\usepackage{multirow}
\usepackage[symbol]{footmisc}

\usepackage[normalem]{ulem}

\newcommand\js{\bgroup\markoverwith{\textcolor[rgb]{0.8, .3, .1}{\rule[0.5ex]{8pt}{1.5pt}}}\ULon}
\newcommand\xyzs{\bgroup\markoverwith{\textcolor[rgb]{0, 0, 1}{\rule[0.5ex]{8pt}{1.5pt}}}\ULon}

\title{Can Language Model Understand Word Semantics as A Chatbot? An Empirical Study of Language Model Internal External Mismatch}

  \author{Jinman Zhao$^{1*}$, Xueyan Zhang$^{2*}$, Xingyu Yue$^{1}$, Weizhe Chen$^{1}$, Zifan Qian$^{3}$, Ruiyu Wang$^{1}$ \\
  $^1$University of Toronto,$^2$Waterloo University,$^3$University of Alberta\\
  jzhao@cs.toronto.edu, xueyan.zhang@uwaterloo.ca
}

\begin{document}
\maketitle
\begin{abstract}
Current common interactions with language models is through full inference. This approach may not necessarily align with the model's internal knowledge. Studies show discrepancies between prompts and internal representations. Most focus on sentences understanding. We study the discrepancy of word semantics understanding in internal and external mismatch across Encoder-only, Decoder-only, and Encoder-Decoder pre-trained language models.\footnote[0]{$^{*}$Equal contribution}
\end{abstract}

\section{Introduction}

Language models (LMs)~\citep{bert,radford2019language,gpt-j,gpt3} have drawn a wide range of interest in many fields.
The ability to process natural language, encode data into parameters, and generate convincing paragraphs drives many people to consider it as trusted knowledge source.
LM's truthfulness is then a key factor in determining if they are suitable for many downstream applications;
in other words, researchers need to assess LMs integrity in their claims.

Machine honesty is very important in recent LLM research. Honesty intersects with aspects such as truthfulness~\citep{evans2021truthful}, calibration~\citep{guo2017calibration,calirevisit,10.1162/tacl_a_00494}, self-knowledge~\citep{llmknwodk,llmknow}, non-deceptiveness~\citep{internallie} and so on. 
There are works investigating whether AI models are aware of what they are expressing. 
The comprehensive analysis on the honesty of LLMs by ~\citet{llmknow} concludes that LLMs are well-calibrated.~\citet{cheng2024can} has similar conclusions regarding models' awareness and understanding of what they know and what they do not know. Other works also demonstrated quirky behaviors and phenomena associated with how the model respond to prompt~\citep{khashabi-etal-2022-prompt,webson-etal-2023-language}. 

 Prior works keep demonstrate that there is a discrepancy between internal and external representations. ~\citet{prompt-prob} explored the discrepancies between the model's internal next token distribution and the distribution obtained using prompts such as \textit{"What is the best next word?"}.
~\citet{probing-query} analyzed the internal and external inconsistencies of the model from the perspectives of probing(internal) and querying(external). ~\citet{internallie} investigated how to use the internal state to determine the truthfulness of text generated by language models, thereby also confirming inconsistencies between the model's internal and external outputs.

In this work, external output refers to the results produced by LMs, specifically the distributions over special positional tokens
(e.g. [MASK] token in Encoder-based LMs, next token in Decoder-based LMs).
Researches show that there are information stores in the internal hidden representation. We use hidden representation as the internal information~\citep{wang2023knowledgeeditinglargelanguage}.
ELMo~\citep{elmo} is the first to introduce the concept of contextual embeddings by adapting embeddings to word usage in context. Before that word embeddings are static~\cite{word2vec,glove}.
BERT~\citep{bert} utilizes transformer architecture to capture deep contextual nuances, setting new standards for various tasks. 

Word embeddings represent the contextual meaning of a word using high-dimensional vectors. In this work, we employed probes and queries to compare language models across three commonly used word embedding evaluation benchmarks. 
Previous research by \citet{probing-query} found no significant difference between queries and probes in question-answering tasks, which primarily focus on sentence-level meaning extraction. 
However, our results diverge markedly from these findings; we observed a substantial gap between probes and queries, highlighting potential limitations of queries in capturing word-level semantics.

\section{Method}

To investigate LM's understanding on word semantics,
we mainly focus on 3 distinct tasks spanning the spectrum of LM training streams;
namely word similarity, structured prediction, and Analogy.
First, we introduce the benchmark, followed by the strategy of \textit{probing} and \textit{querying}. 

We employ the linear probing, which are commonly used in recent NLP works~\citep{probing-query,marks2024geometrytruthemergentlinear}. Compared to the finetuning process, linear probing takes only thousands of parameters which is significantly smaller than the LMs itself with millions to billions of parameters.

\subsection{Word Similarity}
\paragraph{Benchmark} Word similarity tasks~\citep{wordsim353,RW} are used to test semantic similarities between words. We use WiC~\citep{wic} to test the similarity of contextual embedding. WiC contains 5428 test data and 1400 training data. Each data contains a pair of sentences that both contain the target word, and the golden is to answer whether the target word in two sentences has the same meaning contextually.

\paragraph{Probe} Let $t_1,...t_m$ be the tokens that construct the target sentence. $\vec{h_{1,1}},..\vec{h_{1,m}}$ be the hidden vector of target word tokens in the first sentence. We use the average vector $\vec{h_1} =\frac{1}{m}\sum_{i}{\vec{h_{1,i}}}$ to represent the target word in the first sentence.

Similarly, we use $\vec{h_2}$ to represent the target word in the other sentence. We adopt the classification objective function in ~\cite{sbert} that takes $[\vec{h1}, \vec{h2}, |\vec{h1}-\vec{h2}|]$ as input and build a 2-class logistic regression on top:
$$out = Softmax(Linear([\vec{h1}, \vec{h2}, |\vec{h1}-\vec{h2}|]))$$
\paragraph{Query}
We use the queries that are commonly used in other work~\cite{wei2022finetuned}. For example:
\begin{framed}
\noindent\textit{\{Sentence1\}\\\{Sentence2\}\\ Does the word "\{word\}" mean the same thing in the above two sentences?\\
Answer:[MASK]}
\end{framed}

The prompts we used are listed in Appendix~\ref{sec:WICprompt}. We report the accuracy with the highest accuracy. For generative LMs, we will ask LMs to generate [MASK] position tokens.
After the inference, we extract the result logits and compare the probability of the expected output token; 
for example, Bert is expected to output token 'Yes' or 'No', and then a normalized probability is computed.

\subsection{Structured Prediction}
\paragraph{Benchmark}Named Entity Recognition (NER) ~\citep{conll2003,WNUT-17} task is to identify and classify entities (like names of persons, organizations, locations and etc.) in a given text. NER is also used to evaluate word embeddings~\citep{glove}. In this work, we use CoNLL2003~\citep{conll2003} which contains 46,435 tokens in the test set. CoNLL2003 has four entities: person, location, miscellaneous and organization. Detailed statistics are listed in Appendix~\ref{sec:conll2003stats}. 

\paragraph{Probe} Similarly, we use $\vec{h_i}$ to be the average hidden vectors of all tokens in the $i^{th}$ word. We then build a 5-class logistic regression:
$$out = Softmax(Linear(\vec{h_i}))$$


\paragraph{Query} 
After comparing the accuracy of many prompts, we adopt the following:
\begin{framed}
    \noindent\{\textit{Sentence}\}. The word \{\textit{word}\} in the previous sentence is labelled as [MASK]
\end{framed}
\noindent We compare the probability of \textit{``location'', ``person'', ``organization'', ``miscellaneous''} and select the one with the highest score as the output.
\subsection{Analogy}
\paragraph{Benchmark} BATS~\citep{BATS} is an analogy dataset containing 199 validation data and 1799 test data. BATS is commonly used to evaluate the quality of word embeddings by testing their ability to capture semantic and syntactic relationships between words. This benchmark contains multiple-choice questions that give stem words \textit{a} and \textit{b} and ask to choose the best pair of words from 4 choices that best fit \textit{" a is to b as c is to d?"}.
For example, given the stem pairs \textit{("einstein", "physicist")} and 4 choices pairs \textit{("bee", "larva"), ("schwarzenegger", "napoleon"), ("pascal", "mathematician"), ("locke", "Confucius")}, apparently the pair \textit{("pascal", "mathematician")} should be chosen since it has the closest relation as the stem pair.
\paragraph{Probe} We first use GPT-4 to generate 5 sentences for each word in the BATS. Then compute hidden vectors of each word of each sentence. Then average 5 word vectors to be the vector representation of each word. For the probe, each data has three negative samples and one positive sample, which makes the training data unbalanced. We follow ~\citep{ushio-etal-2021-bert}, for gold analogies, we put both (a, b)-(c, d) and (a, c) - (b, d) as positive samples. This would increase the size of the positive samples. Let $\vec{h_a}$ be the vector representation of word $a$ and so on. For the analogy question, the distance from \textit{b} to \textit{a} should be similar to the distance from \textit{d} and \textit{c}. Therefore we also inherit classification objection~\cite{sbert}.
\begin{align}
    \vec{c} = |[\vec{h_a}-\vec{h_b}&,\vec{h_c}-\vec{h_d},\vec{h_a}-\vec{h_b}+\vec{h_d}-\vec{h_c}]|\\
    out &= Softmax(Linear(\vec{c}))
\end{align}
During the evaluation step, the pair with the highest positive probability will be chosen.
\paragraph{Query}
We select the following prompt:
\begin{framed}
    \noindent
    \textit{\{$Stem_1$\} is to \{$Stem_2$\} as:\\
    A) \{$Choice_1a$\} is to \{$Choice_1b$\} \\
    B) \{$Choice_2a$\} is to \{$Choice_2b$\} \\
    C) \{$Choice_3a$\} is to \{$Choice_3b$\} \\
    D) \{$Choice_4a$\} is to \{$Choice_4b$\} \\
    Answer:[MASK]}
\end{framed}
\noindent Other prompts are listed in Appendix~\ref{sec:Anaprompts}.

\section{Results}
\paragraph{Model Selection}
We selected three fundamentally different language models based on the architecture.
\begin{itemize}
    \item For the encoder-based model, we choose BERT model~\citep{devlin-etal-2019-bert}. 
    \item For the decoder-based models, we opt for both GPT-2~\citep{radford2019language}.
    \item  For the encoder-decoder-based model evaluation, we select T5~\citep{raffel2023exploringlimitstransferlearning}. 
\end{itemize}

\subsection{Main Results}

Table~\ref{tbl:main} shows the accuracy achieved by representative models in the target benchmark. We found noticeable differences between probe and query in terms of word semantic capturing. This gap is evident across all models and all benchmarks, highlighting that pretrained language models, when used as chatbots, can exhibit information discrepancies compared to the knowledge stored within their internal neurons.

In WiC benchmark, the answer to the prompt question is binary (yes-no question);
we observe that all models are query accuracy is within the range of 49\% to 53\%, close to random guess (50\%).
Probe accuracy is considerably higher with a highest 65\% chance to correctly understand context-sentence word semantics.
As aforementioned, because probing performs linear classification directly on the word embedding, the higher accuracy above random guess indicate that the internal representation is indeed capable to distinguish the word similarity; however, this knowledge failed to propagate to the model output.

F1 score is a common indicator for NER tasks;
we observed a more pronounced internal-external discrepancy. 
Because models with encoder have a better understanding of the input words, they outperform decoder-only models.
For instance, BERT embeddings for probing achieved state-of-the-art performance with an F1 score of 96\%.
GPT-2, on the other hand, has a much lower F1 score, conforming to the observation made by~\citet{wang2023gptnernamedentityrecognition} and~\citet{xie-etal-2023-empirical}, where GPT3/ChatGPT in both fine-tune and zero-shot setting is less performant than BERT. 
In contrast, the performance of queries was even lower than random guessing.

Given that the prompt in Analogy benchmark is a multiple choice question with four options,
BERT models exhibits a nearly random guess accuracy around 25\% in query, while the probe accuracy almost doubles.
The query accuracy of GPT and T5 models direct some of their understanding to the output, reaching around 30\%.
GPT-2 has the lowest probe accuracy at 41\%; it may reflect that decoder-based models are more suitable for text generation and less performant in extracting the meaning of words.

\begin{table*}[h]
\centering
\begin{tabular}{cc|c|ccc|c}
\hline
\multirow{2}{*}{Model}    &  \multirow{2}{*}{method}     & WiC  & \multicolumn{3}{c|}{NER}  & Analogy \\ \cline{3-7} 
        &       &  Acc(\%)    & Precision & Recall & F1  & Acc(\%)     \\ \hline
\multirow{2}{*}{BERT-base}    & Query & 50 & 7 & 100 & 14& 25   \\
        & Probe & 65 & 95 & 96 & 96& 51     \\   \hline
\multirow{2}{*}{BERT-large}    & Query & 53 & 3 & 100 & 6& 26   \\
        & Probe & 65 &  96& 95 &96 & 48    \\ \hline
\multirow{2}{*}{GPT-2}    & Query & 49 & 4& 42 & 8& 33    \\
        & Probe & 58 &97& 32& 48& 41    \\ \hline        
\multirow{2}{*}{T5-small}& Query & 49 & 5 & 8& 6& 31    \\
        & Probe & 61 & 98& 94& 96& 47    \\ \hline
\multirow{2}{*}{T5-large}& Query & 50 & 4& 6& 5&  35   \\
        & Probe & 65 & 99& 96& 97& 48    \\ \hline
\end{tabular}
\caption{Accuracy of encoder, decoder, and encoder-decoder models on benchmark WIC, NER, and Analogy.}
\label{tbl:main}
\end{table*}




\subsection{Instruct Tuning and Finetuning}

When there is a mismatch between internal and external representation,
it may indicate an alignment issue; 
the knowledge of the model is not properly propagated to the very end.
We then investigate if finetuning improves the misalignment issue.

Flan T5 is a instruction-finetune model based on T5 in a mixture of tasks~\cite{raffel2023exploringlimitstransferlearning,wei2022finetuned}; specifically, WiC is explicitly used as one of the datasets.
As shown in Table~\ref{tab:flan},
Flan T5 outperforms the T5 in terms of query accuracy, proving that finetuning indeed enhances model's ability to direct the knowledge to the output.
A similar observation can be found in \citet{probing-query},
where the authors finetune GPT2-XL on true question/answer pairs.
However, although the accuracy is boosted from 50\% to 59\%, probing still shows a better performance.
The model seems to have a similar understanding of word semantics in both models, and thus Flan T5 slightly improves probe accuracy from 65\% to 68\% compared to T5.

\begin{table}[ht]
    \centering
    \begin{tabular}{c|c|c}
    \hline
    Model         & Method & WiC \\\hline
    T5-large      & Query  & 50  \\
                  & Probe  & 65  \\\hline
    Flan-T5-large & Query  & 59  \\
                  & Probe  & 68 \\\hline
    \end{tabular}
    \caption{Accuracy of T5 and Flan-T5.}
    \label{tab:flan}
\end{table}


\subsection{Calibration}
\begin{figure}[h!]
    \centering
    \begin{tikzpicture}
        \node[rotate=90] at (-0.5, 7) {\textbf{BERT-base}};
        \node[rotate=90] at (-0.5, 4) {\textbf{GPT-2}};
        \node[rotate=90] at (-0.5, 1) {\textbf{T5-large}};

        \node[rotate=0] at (1.5, 9) {\textbf{Probe}};
        \node[rotate=0] at (5, 9) {\textbf{Query}};

        \node at (1.5, 7) {\includegraphics[width=3cm]{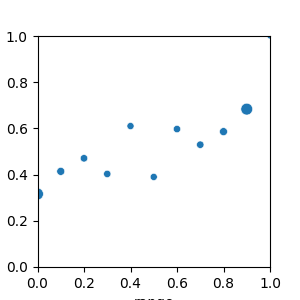}};
        \node at (1.5, 4) {\includegraphics[width=3cm]{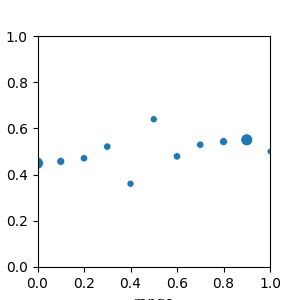}};
        \node at (1.5, 1) {\includegraphics[width=3cm]{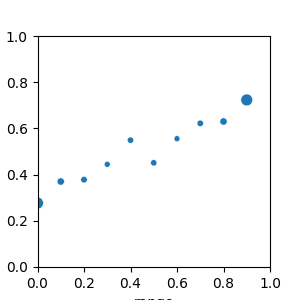}};

        \node at (5, 7) {\includegraphics[width=3cm]{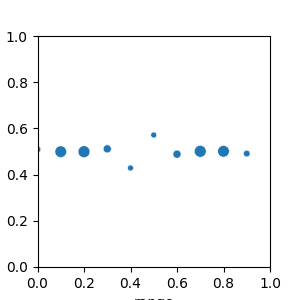}};
        \node at (5, 4) {\includegraphics[width=3cm]{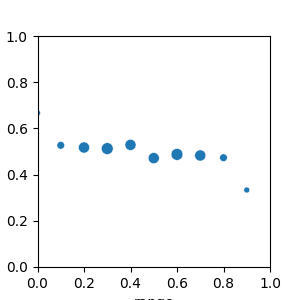}};
        \node at (5, 1) {\includegraphics[width=3cm]{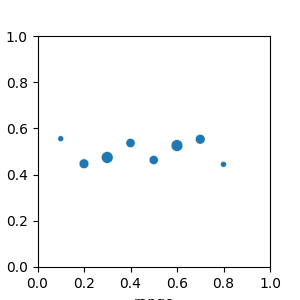}};

        \node at (3.25, -1) {\text{Confidence}};
        \node[rotate=90] at (6.5, 4) {\text{Accuracy}};
    \end{tikzpicture}
    \caption{Model confidence and Accuracy comparison on WiC datasets.}\label{calibration}
\end{figure}
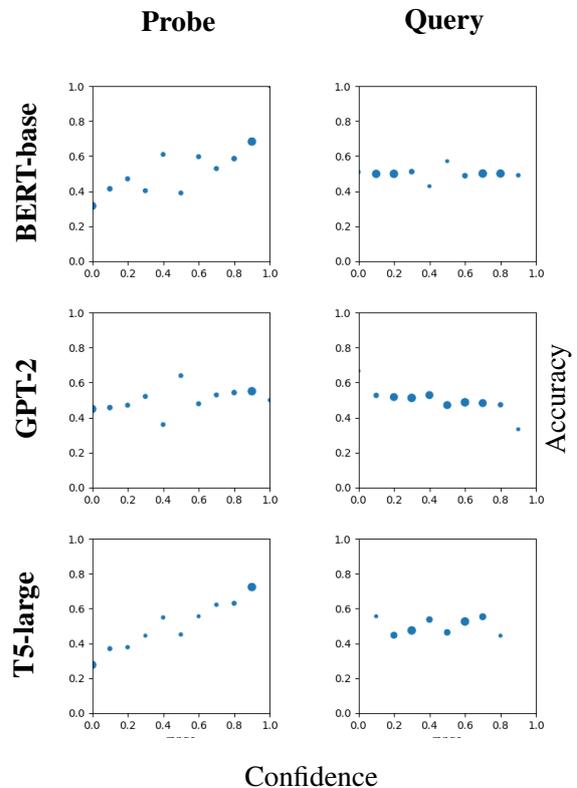

A well-calibrated model should exhibit close alignment between confidence and accuracy. We demonstrate the confidence and accuracy of three models on the WIC task in Figure~\ref{calibration}; probe are better calibrated than queries. Furthermore, model with better WiC performance like BERT and T5 has the best calibration than GPT-2.

\section{Conclusion}
In this paper, we studied the discrepancy between language model's internal and external representations. We mainly focus on the ability to understand the word semantics.
Probe consistently shows a better performance than query, indicating that there is potential to improve models truthfulness. Currently, the model knowledge is not properly reflected on the model's generated output. We find that finetuning or calibration help to improve the accuracy to some extend, but it still not on par to probe accuracy. Other factors like model size also contribute to the discrepancy. Improving the model's truthfulness will unleash their potential in applications where reliability and robustness are preferable.

\section*{Limitation}
Due to limitations in hardware resources and budget constraints, the number of models included in our study is relatively limited. Although we selected representative models to validate our hypotheses, this limitation might affect the generalizability of our findings. Additionally, with restricted computational capacity, we were unable to explore more complex model architectures, which could have provided deeper insights into specific issues. Future research could expand the scope of model selection and explore more diverse and intricate models by securing additional resources, thus enhancing the comprehensiveness and accuracy of the study.

\begin{thebibliography}{34}
\providecommand{\natexlab}[1]{#1}

\bibitem[{Azaria and Mitchell(2023)}]{internallie}
Amos Azaria and Tom Mitchell. 2023.
\newblock \href {https://doi.org/10.18653/v1/2023.findings-emnlp.68} {The internal state of an {LLM} knows when it{'}s lying}.
\newblock In \emph{Findings of the Association for Computational Linguistics: EMNLP 2023}, pages 967--976, Singapore. Association for Computational Linguistics.

\bibitem[{Brown et~al.(2020)Brown, Mann, Ryder, Subbiah, Kaplan, Dhariwal, Neelakantan, Shyam, Sastry, Askell, Agarwal, Herbert-Voss, Krueger, Henighan, Child, Ramesh, Ziegler, Wu, Winter, Hesse, Chen, Sigler, Litwin, Gray, Chess, Clark, Berner, McCandlish, Radford, Sutskever, and Amodei}]{gpt3}
Tom Brown, Benjamin Mann, Nick Ryder, Melanie Subbiah, Jared~D Kaplan, Prafulla Dhariwal, Arvind Neelakantan, Pranav Shyam, Girish Sastry, Amanda Askell, Sandhini Agarwal, Ariel Herbert-Voss, Gretchen Krueger, Tom Henighan, Rewon Child, Aditya Ramesh, Daniel Ziegler, Jeffrey Wu, Clemens Winter, Chris Hesse, Mark Chen, Eric Sigler, Mateusz Litwin, Scott Gray, Benjamin Chess, Jack Clark, Christopher Berner, Sam McCandlish, Alec Radford, Ilya Sutskever, and Dario Amodei. 2020.
\newblock \href {https://proceedings.neurips.cc/paper_files/paper/2020/file/1457c0d6bfcb4967418bfb8ac142f64a-Paper.pdf} {Language models are few-shot learners}.
\newblock In \emph{Advances in Neural Information Processing Systems}, volume~33, pages 1877--1901. Curran Associates, Inc.

\bibitem[{Cheng et~al.(2024)Cheng, Sun, Liu, Zhang, Yin, Li, Li, Chen, and Qiu}]{cheng2024can}
Qinyuan Cheng, Tianxiang Sun, Xiangyang Liu, Wenwei Zhang, Zhangyue Yin, Shimin Li, Linyang Li, Kai Chen, and Xipeng Qiu. 2024.
\newblock Can ai assistants know what they don't know?
\newblock \emph{arXiv preprint arXiv:2401.13275}.

\bibitem[{Derczynski et~al.(2017)Derczynski, Nichols, van Erp, and Limsopatham}]{WNUT-17}
Leon Derczynski, Eric Nichols, Marieke van Erp, and Nut Limsopatham. 2017.
\newblock \href {https://doi.org/10.18653/v1/W17-4418} {Results of the {WNUT}2017 shared task on novel and emerging entity recognition}.
\newblock In \emph{Proceedings of the 3rd Workshop on Noisy User-generated Text}, pages 140--147, Copenhagen, Denmark. Association for Computational Linguistics.

\bibitem[{Devlin et~al.(2019{\natexlab{a}})Devlin, Chang, Lee, and Toutanova}]{bert}
Jacob Devlin, Ming-Wei Chang, Kenton Lee, and Kristina Toutanova. 2019{\natexlab{a}}.
\newblock \href {https://doi.org/10.18653/v1/N19-1423} {{BERT}: Pre-training of deep bidirectional transformers for language understanding}.
\newblock In \emph{Proceedings of the 2019 Conference of the North {A}merican Chapter of the Association for Computational Linguistics: Human Language Technologies, Volume 1 (Long and Short Papers)}, pages 4171--4186, Minneapolis, Minnesota. Association for Computational Linguistics.

\bibitem[{Devlin et~al.(2019{\natexlab{b}})Devlin, Chang, Lee, and Toutanova}]{devlin-etal-2019-bert}
Jacob Devlin, Ming-Wei Chang, Kenton Lee, and Kristina Toutanova. 2019{\natexlab{b}}.
\newblock \href {https://doi.org/10.18653/v1/N19-1423} {{BERT}: Pre-training of deep bidirectional transformers for language understanding}.
\newblock In \emph{Proceedings of the 2019 Conference of the North {A}merican Chapter of the Association for Computational Linguistics: Human Language Technologies, Volume 1 (Long and Short Papers)}, pages 4171--4186, Minneapolis, Minnesota. Association for Computational Linguistics.

\bibitem[{Evans et~al.(2021)Evans, Cotton-Barratt, Finnveden, Bales, Balwit, Wills, Righetti, and Saunders}]{evans2021truthful}
Owain Evans, Owen Cotton-Barratt, Lukas Finnveden, Adam Bales, Avital Balwit, Peter Wills, Luca Righetti, and William Saunders. 2021.
\newblock Truthful ai: Developing and governing ai that does not lie.
\newblock \emph{arXiv preprint arXiv:2110.06674}.

\bibitem[{Finkelstein et~al.(2001)Finkelstein, Gabrilovich, Matias, Rivlin, Solan, Wolfman, and Ruppin}]{wordsim353}
Lev Finkelstein, Evgeniy Gabrilovich, Yossi Matias, Ehud Rivlin, Zach Solan, Gadi Wolfman, and Eytan Ruppin. 2001.
\newblock Placing search in context: The concept revisited.
\newblock In \emph{Proceedings of the 10th international conference on World Wide Web}, pages 406--414.

\bibitem[{Gladkova et~al.()Gladkova, Drozd, and Matsuoka}]{BATS}
Anna Gladkova, Aleksandr Drozd, and Satoshi Matsuoka.
\newblock Analogy-based detection of morphological and semantic relations with word embeddings: What works and what doesn't.
\newblock In \emph{Proceedings of the NAACL-HLT SRW, address = {San Diego, California, June 12-17, 2016}, publisher = {ACL}, year = {2016}, pages = {47-54} doi = {10.18653/v1/N16-2002}, url = {https://www.aclweb.org/anthology/N/N16/N16-2002.pdf},}.

\bibitem[{Guo et~al.(2017)Guo, Pleiss, Sun, and Weinberger}]{guo2017calibration}
Chuan Guo, Geoff Pleiss, Yu~Sun, and Kilian~Q Weinberger. 2017.
\newblock On calibration of modern neural networks.
\newblock In \emph{International conference on machine learning}, pages 1321--1330. PMLR.

\bibitem[{Hu and Levy(2023)}]{prompt-prob}
Jennifer Hu and Roger Levy. 2023.
\newblock \href {https://doi.org/10.18653/v1/2023.emnlp-main.306} {Prompting is not a substitute for probability measurements in large language models}.
\newblock In \emph{Proceedings of the 2023 Conference on Empirical Methods in Natural Language Processing}, pages 5040--5060, Singapore. Association for Computational Linguistics.

\bibitem[{Kadavath et~al.(2022)Kadavath, Conerly, Askell, Henighan, Drain, Perez, Schiefer, Hatfield-Dodds, DasSarma, Tran-Johnson et~al.}]{llmknow}
Saurav Kadavath, Tom Conerly, Amanda Askell, Tom Henighan, Dawn Drain, Ethan Perez, Nicholas Schiefer, Zac Hatfield-Dodds, Nova DasSarma, Eli Tran-Johnson, et~al. 2022.
\newblock Language models (mostly) know what they know.
\newblock \emph{arXiv preprint arXiv:2207.05221}.

\bibitem[{Khashabi et~al.(2022)Khashabi, Lyu, Min, Qin, Richardson, Welleck, Hajishirzi, Khot, Sabharwal, Singh, and Choi}]{khashabi-etal-2022-prompt}
Daniel Khashabi, Xinxi Lyu, Sewon Min, Lianhui Qin, Kyle Richardson, Sean Welleck, Hannaneh Hajishirzi, Tushar Khot, Ashish Sabharwal, Sameer Singh, and Yejin Choi. 2022.
\newblock \href {https://doi.org/10.18653/v1/2022.naacl-main.266} {Prompt waywardness: The curious case of discretized interpretation of continuous prompts}.
\newblock In \emph{Proceedings of the 2022 Conference of the North American Chapter of the Association for Computational Linguistics: Human Language Technologies}, pages 3631--3643, Seattle, United States. Association for Computational Linguistics.

\bibitem[{Liu et~al.(2023)Liu, Casper, Hadfield-Menell, and Andreas}]{probing-query}
Kevin Liu, Stephen Casper, Dylan Hadfield-Menell, and Jacob Andreas. 2023.
\newblock \href {https://doi.org/10.18653/v1/2023.emnlp-main.291} {Cognitive dissonance: Why do language model outputs disagree with internal representations of truthfulness?}
\newblock In \emph{Proceedings of the 2023 Conference on Empirical Methods in Natural Language Processing}, pages 4791--4797, Singapore. Association for Computational Linguistics.

\bibitem[{Luong et~al.(2013)Luong, Socher, and Manning}]{RW}
Thang Luong, Richard Socher, and Christopher Manning. 2013.
\newblock \href {https://aclanthology.org/W13-3512} {Better word representations with recursive neural networks for morphology}.
\newblock In \emph{Proceedings of the Seventeenth Conference on Computational Natural Language Learning}, pages 104--113, Sofia, Bulgaria. Association for Computational Linguistics.

\bibitem[{Marks and Tegmark(2024)}]{marks2024geometrytruthemergentlinear}
Samuel Marks and Max Tegmark. 2024.
\newblock \href {https://arxiv.org/abs/2310.06824} {The geometry of truth: Emergent linear structure in large language model representations of true/false datasets}.
\newblock \emph{Preprint}, arXiv:2310.06824.

\bibitem[{Mielke et~al.(2022)Mielke, Szlam, Dinan, and Boureau}]{10.1162/tacl_a_00494}
Sabrina~J. Mielke, Arthur Szlam, Emily Dinan, and Y-Lan Boureau. 2022.
\newblock \href {https://doi.org/10.1162/tacl_a_00494} {{Reducing Conversational Agents’ Overconfidence Through Linguistic Calibration}}.
\newblock \emph{Transactions of the Association for Computational Linguistics}, 10:857--872.

\bibitem[{Mikolov et~al.(2013)Mikolov, Sutskever, Chen, Corrado, and Dean}]{word2vec}
Tomas Mikolov, Ilya Sutskever, Kai Chen, Greg~S Corrado, and Jeff Dean. 2013.
\newblock \href {https://proceedings.neurips.cc/paper_files/paper/2013/file/9aa42b31882ec039965f3c4923ce901b-Paper.pdf} {Distributed representations of words and phrases and their compositionality}.
\newblock In \emph{Advances in Neural Information Processing Systems}, volume~26. Curran Associates, Inc.

\bibitem[{Minderer et~al.(2021)Minderer, Djolonga, Romijnders, Hubis, Zhai, Houlsby, Tran, and Lucic}]{calirevisit}
Matthias Minderer, Josip Djolonga, Rob Romijnders, Frances Hubis, Xiaohua Zhai, Neil Houlsby, Dustin Tran, and Mario Lucic. 2021.
\newblock Revisiting the calibration of modern neural networks.
\newblock \emph{Advances in Neural Information Processing Systems}, 34:15682--15694.

\bibitem[{Pennington et~al.(2014)Pennington, Socher, and Manning}]{glove}
Jeffrey Pennington, Richard Socher, and Christopher Manning. 2014.
\newblock \href {https://doi.org/10.3115/v1/D14-1162} {{G}lo{V}e: Global vectors for word representation}.
\newblock In \emph{Proceedings of the 2014 Conference on Empirical Methods in Natural Language Processing ({EMNLP})}, pages 1532--1543, Doha, Qatar. Association for Computational Linguistics.

\bibitem[{Peters et~al.(2018)Peters, Neumann, Iyyer, Gardner, Clark, Lee, and Zettlemoyer}]{elmo}
Matthew~E. Peters, Mark Neumann, Mohit Iyyer, Matt Gardner, Christopher Clark, Kenton Lee, and Luke Zettlemoyer. 2018.
\newblock \href {https://doi.org/10.18653/v1/N18-1202} {Deep contextualized word representations}.
\newblock In \emph{Proceedings of the 2018 Conference of the North {A}merican Chapter of the Association for Computational Linguistics: Human Language Technologies, Volume 1 (Long Papers)}, pages 2227--2237, New Orleans, Louisiana. Association for Computational Linguistics.

\bibitem[{Pilehvar and Camacho-Collados(2019)}]{wic}
Mohammad~Taher Pilehvar and Jose Camacho-Collados. 2019.
\newblock \href {https://doi.org/10.18653/v1/N19-1128} {{W}i{C}: the word-in-context dataset for evaluating context-sensitive meaning representations}.
\newblock In \emph{Proceedings of the 2019 Conference of the North {A}merican Chapter of the Association for Computational Linguistics: Human Language Technologies, Volume 1 (Long and Short Papers)}, pages 1267--1273, Minneapolis, Minnesota. Association for Computational Linguistics.

\bibitem[{Radford et~al.(2019)Radford, Wu, Child, Luan, Amodei, and Sutskever}]{radford2019language}
Alec Radford, Jeff Wu, Rewon Child, David Luan, Dario Amodei, and Ilya Sutskever. 2019.
\newblock Language models are unsupervised multitask learners.

\bibitem[{Raffel et~al.(2023)Raffel, Shazeer, Roberts, Lee, Narang, Matena, Zhou, Li, and Liu}]{raffel2023exploringlimitstransferlearning}
Colin Raffel, Noam Shazeer, Adam Roberts, Katherine Lee, Sharan Narang, Michael Matena, Yanqi Zhou, Wei Li, and Peter~J. Liu. 2023.
\newblock \href {https://arxiv.org/abs/1910.10683} {Exploring the limits of transfer learning with a unified text-to-text transformer}.
\newblock \emph{Preprint}, arXiv:1910.10683.

\bibitem[{Reimers and Gurevych(2019)}]{sbert}
Nils Reimers and Iryna Gurevych. 2019.
\newblock \href {https://doi.org/10.18653/v1/D19-1410} {Sentence-{BERT}: Sentence embeddings using {S}iamese {BERT}-networks}.
\newblock In \emph{Proceedings of the 2019 Conference on Empirical Methods in Natural Language Processing and the 9th International Joint Conference on Natural Language Processing (EMNLP-IJCNLP)}, pages 3982--3992, Hong Kong, China. Association for Computational Linguistics.

\bibitem[{Tjong Kim~Sang and De~Meulder(2003)}]{conll2003}
Erik~F. Tjong Kim~Sang and Fien De~Meulder. 2003.
\newblock \href {https://aclanthology.org/W03-0419} {Introduction to the {C}o{NLL}-2003 shared task: Language-independent named entity recognition}.
\newblock In \emph{Proceedings of the Seventh Conference on Natural Language Learning at {HLT}-{NAACL} 2003}, pages 142--147.

\bibitem[{Ushio et~al.(2021)Ushio, Espinosa~Anke, Schockaert, and Camacho-Collados}]{ushio-etal-2021-bert}
Asahi Ushio, Luis Espinosa~Anke, Steven Schockaert, and Jose Camacho-Collados. 2021.
\newblock \href {https://doi.org/10.18653/v1/2021.acl-long.280} {{BERT} is to {NLP} what {A}lex{N}et is to {CV}: Can pre-trained language models identify analogies?}
\newblock In \emph{Proceedings of the 59th Annual Meeting of the Association for Computational Linguistics and the 11th International Joint Conference on Natural Language Processing (Volume 1: Long Papers)}, pages 3609--3624, Online. Association for Computational Linguistics.

\bibitem[{Wang and Komatsuzaki(2021)}]{gpt-j}
Ben Wang and Aran Komatsuzaki. 2021.
\newblock {GPT-J-6B: A 6 Billion Parameter Autoregressive Language Model}.
\newblock \url{https://github.com/kingoflolz/mesh-transformer-jax}.

\bibitem[{Wang et~al.(2023{\natexlab{a}})Wang, Sun, Li, Ouyang, Wu, Zhang, Li, and Wang}]{wang2023gptnernamedentityrecognition}
Shuhe Wang, Xiaofei Sun, Xiaoya Li, Rongbin Ouyang, Fei Wu, Tianwei Zhang, Jiwei Li, and Guoyin Wang. 2023{\natexlab{a}}.
\newblock \href {https://arxiv.org/abs/2304.10428} {Gpt-ner: Named entity recognition via large language models}.
\newblock \emph{Preprint}, arXiv:2304.10428.

\bibitem[{Wang et~al.(2023{\natexlab{b}})Wang, Zhu, Liu, Zheng, Chen, and Li}]{wang2023knowledgeeditinglargelanguage}
Song Wang, Yaochen Zhu, Haochen Liu, Zaiyi Zheng, Chen Chen, and Jundong Li. 2023{\natexlab{b}}.
\newblock \href {https://arxiv.org/abs/2310.16218} {Knowledge editing for large language models: A survey}.
\newblock \emph{Preprint}, arXiv:2310.16218.

\bibitem[{Webson et~al.(2023)Webson, Loo, Yu, and Pavlick}]{webson-etal-2023-language}
Albert Webson, Alyssa Loo, Qinan Yu, and Ellie Pavlick. 2023.
\newblock \href {https://doi.org/10.18653/v1/2023.findings-emnlp.514} {Are language models worse than humans at following prompts? it{'}s complicated}.
\newblock In \emph{Findings of the Association for Computational Linguistics: EMNLP 2023}, pages 7662--7686, Singapore. Association for Computational Linguistics.

\bibitem[{Wei et~al.(2022)Wei, Bosma, Zhao, Guu, Yu, Lester, Du, Dai, and Le}]{wei2022finetuned}
Jason Wei, Maarten Bosma, Vincent Zhao, Kelvin Guu, Adams~Wei Yu, Brian Lester, Nan Du, Andrew~M. Dai, and Quoc~V Le. 2022.
\newblock \href {https://openreview.net/forum?id=gEZrGCozdqR} {Finetuned language models are zero-shot learners}.
\newblock In \emph{International Conference on Learning Representations}.

\bibitem[{Xie et~al.(2023)Xie, Li, Zhang, Zhang, Liu, and Wang}]{xie-etal-2023-empirical}
Tingyu Xie, Qi~Li, Jian Zhang, Yan Zhang, Zuozhu Liu, and Hongwei Wang. 2023.
\newblock \href {https://doi.org/10.18653/v1/2023.emnlp-main.493} {Empirical study of zero-shot {NER} with {C}hat{GPT}}.
\newblock In \emph{Proceedings of the 2023 Conference on Empirical Methods in Natural Language Processing}, pages 7935--7956, Singapore. Association for Computational Linguistics.

\bibitem[{Yin et~al.(2023)Yin, Sun, Guo, Wu, Qiu, and Huang}]{llmknwodk}
Zhangyue Yin, Qiushi Sun, Qipeng Guo, Jiawen Wu, Xipeng Qiu, and Xuanjing Huang. 2023.
\newblock \href {https://doi.org/10.18653/v1/2023.findings-acl.551} {Do large language models know what they don{'}t know?}
\newblock In \emph{Findings of the Association for Computational Linguistics: ACL 2023}, pages 8653--8665, Toronto, Canada. Association for Computational Linguistics.

\end{thebibliography}

\appendix
\section{WiC Prompt}
\label{sec:WICprompt}
See Table~\ref{tab:wicprompts} for the list of prompts we use in WIC evaluation.
\begin{table*}[ht]
\centering
\begin{tabular}{l}
\hline
\textbf{Prompt}  \\
\hline
\{sentence1\}\\\{sentence2\}\\ Does the word "\{word\}" mean the same thing in the above two sentences?\\
Answer:[MASK]  \\
\hline
Sentence 1: \{sentence1\}\\
Sentence 2: \{sentence2\}\\
Does \{word\} mean the same thing in these two sentences?\\
Answer:[MASK]  \\
\hline
Here is one sentence: \{sentence1\}\\
Here is another sentence: \{sentence2\}\\
Does the term \{word\} mean the same thing in both these sentences?\\
Answer:[MASK]  \\
\hline
In these two sentences (1) \{sentence1\} (2) \{sentence2\},\\
does the word \{word\} mean the same thing?\\
Answer:[MASK]  \\
\hline
Does the word "\{word\}" have the same meaning in the following two sentences?\\
\{sentence1\}\\
\{sentence2\}\\
Answer:[MASK]  \\
\hline
Is the word "\{word\}" used in the same way in the following two sentences?\\
\{sentence1\}\\
\{sentence2\}\\
Answer:[MASK]  \\
\hline
Does the word "\{word\}" have the same definition in the next two sentences?\\
\{sentence1\}\\
\{sentence2\}\\
Answer:[MASK]  \\
\hline
Is \{word\} used to mean the same thing in the next two sentences?\\
\{sentence1\}\\
\{sentence2\}\\
Answer:[MASK]  \\
\hline
Does "\{word\}" mean the same thing in these two sentences?\\
\{sentence1\}\\
\{sentence2\}\\
Answer:[MASK]  \\
\hline
Does the word "\{word\}" mean the same thing in "\{sentence1\}" and "\{sentence2\}"?\\
Answer:[MASK]  \\
\hline
\end{tabular}\caption{Prompts for WIC.}\label{tab:wicprompts}
\end{table*}

\section{Analogy Question Prompts}
\label{sec:Anaprompts}
See Table~\ref{tab:analogyquestionprompt} for the prompts we use for analogy question.
\begin{table*}[ht]
\centering
\begin{tabular}{l}
\hline
\textbf{Prompt}  \\
\hline
\{$Stem_1$\} is to \{$Stem_2$\} as:\\
A) \{\} is to \{\} \\
B) \{\} is to \{\} \\
C) \{\} is to \{\} \\
D) \{\} is to \{\} \\
Answer:[MASK]\\
\hline
Which of the following pairs has the most similar relation with \{$Stem_1$, $Stem_2$\}?\\
A) \{, \} \\
B) \{, \} \\
C) \{, \} \\
D) \{, \} \\
Answer:[MASK]  \\
\hline

\end{tabular}\caption{Prompts for Analogy question.}\label{tab:analogyquestionprompt}
\end{table*}
\section{CONLL2003 Statistics}
\label{sec:conll2003stats}
See Table~\ref{tab:conll2003stats} for CoNLL2003 statistics.
\begin{table}[h]
\centering

\begin{tabular}{l|rrr}
\hline
\textbf{Dataset}        & \textbf{Sentences} & \textbf{Tokens} & \textbf{Entities} \\
\hline
Train            & 14,041             & 203,621         & 23,499            \\
Dev         & 3,250              & 51,362          & 5,942             \\
Test                 & 3,453              & 46,435          & 5,648             \\
\hline
\end{tabular}\caption{CoNLL2003 Statistics.}~\label{tab:conll2003stats}
\end{table}

\end{document}